# Gait trajectory generation for a five link bipedal robot based on a reduced dynamical model

[1]*Yosra Arous* & [2]*Olfa Boubaker*
[1,2] National Institute of Applied sciences and Technology,
Tunis, Tunisia
[2]olfa.boubaker@insat.rnu.tn

*Abstract*— **In this paper, a simple trajectory generation method for biped walking is proposed. The dynamic model of the five link bipedal robot is first reduced using several biologically inspired assumptions. A sinusoidal curve is then imposed to the ankle of the swing leg's trajectory. The reduced model is finally obtained and solved: it is an homogeneous 2$^{nd}$ order differential equations with constant coefficients. The algebraic solution obtained ensures a stable rhythmic gait for the bipedal robot. It's continuous in the defined time interval, easy to implement when the boundary conditions are well defined.**

## I. INTRODUCTION

Gait pattern generation [1] is one of key problems of research devoted to bipedal robots. Two kinds of works dedicated to the bipedal walking pattern generation can be distinguished: studies assimilating robots as elementary models and works considering all morphological data of the robot, see [2] and references therein. For the first approach, the linear inverted pendulum model concept is the mostly used concept in order to generate the gait trajectory [3, 4, 5]. For the second group of works attention is paid on the generation of a trajectory tracking control using objective function composed of one or more terms to minimize [2, 6, 7].

In this paper, a simple trajectory generation method is proposed. The dynamic model of the five link bipedal robot is first reduced using several biologically inspired assumptions. A sinusoidal curve is then imposed to the ankle of the swing leg's trajectory. The reduced model is finally obtained and solved: it is an homogeneous 2$^{nd}$ order differential equations with constant coefficients. The solution gives an algebraic solution for joint desired trajectories that ensures a stable rhythmic gait for the bipedal robot.

## II. THE FIVE LINK BIPEDAL ROBOT

The planar bipedal robot prototype is composed of five links associated to five DOF. Fig.1 shows the involved rotations for each link. All physical parameters involved in the kinematic and dynamic models are given by Table 1. They are inspired from [8]. The parameters $m_i, L_i, k_i$ and $I_i$, ($i = 1,..5$), design mass, length, position of center of mass and inertia about center of mass, respectively.

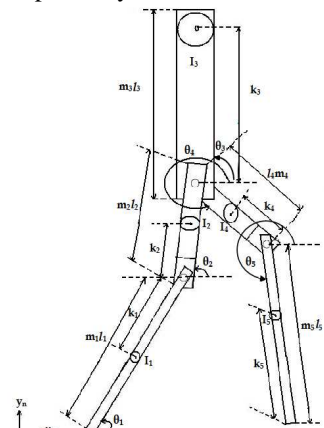

Figure 1. The five link bipedal robot

Table1. Physical parameters of the bipedal robot [8]

| link | Right leg | Right thigh | Pelvis | Left thigh | Left leg |
|---|---|---|---|---|---|
| joint | Right ankle | Right knee | Right hip | Left hip | Left knee |
| Link number | 1 | 2 | 3 | 4 | 5 |
| Mass (kg) | 3.255 | 7.000 | 24.850 | 7.000 | 3.255 |
| Lenght (m) | 0.426 | 0.424 | 0.299 | 0.424 | 0.426 |
| Center of mass (kg.m) | 0.164 | 0.366 | 1.530 | 0.366 | 0.164 |
| Inertia about centre of mass (Kgm$^2$) | 0.184 | 0.184 | 0.206 | 0.184 | 0.184 |

## III. KINEMATIC AND DYNAMIC MODELING

The five-link bipedal robotic system with five degrees of freedom can be described by the following direct kinematic model:

$$X = h(\theta) \quad (1)$$

where $\theta = [\theta_1\ \theta_2\ \theta_3\ \theta_4\ \theta_5]^T \in \Re^5$ is the joint displacement vector, $X = [x\ y]^T \in \Re^2$ is the Cartesian position vector and $h(\theta) \in \Re^2$ is a nonlinear function described by:

$$h(\theta) = \begin{bmatrix} l_1 \cos\theta_1 + l_2 \cos\theta_2 - l_4 \cos\theta_4 - l_5 \cos\theta_5 \\ l_1 \sin\theta_1 + l_2 \sin\theta_2 + l_4 \sin\theta_4 + l_5 \sin\theta_5 \end{bmatrix}$$

The time derivative of the direct kinematic model (1) yields the following differential kinematic model:

$$\dot{X} = J(\theta)\dot{\theta} \quad (2)$$

where $\dot{X} = [\dot{x}\ \dot{y}]^T$ is the Cartesian velocity vector, $\dot{\theta} = [\dot{\theta}_1\ \dot{\theta}_2\ \dot{\theta}_3\ \dot{\theta}_4\ \dot{\theta}_5]^T$ is the vector of joint velocities and $J(\theta)$ is the so-called analytical Jacobian matrix given by:

$$J(\theta) = \begin{pmatrix} -l_1 \sin\theta_1 & -l_2 \sin\theta_2 & 0 & l_4 \sin\theta_4 & l_5 \sin\theta_5 \\ l_1 \cos\theta_1 & l_2 \cos\theta_2 & 0 & l_4 \cos\theta_4 & l_5 \cos\theta_5 \end{pmatrix}$$

A typical walking cycle may include three phases [1]: the single support phase (SSP), the impact phase (IP) and the double support phase (DSP), (see Fig.2). In this paper we focus only on SSP.

Using Lagrange approach [9], the dynamical model of the bipedal robot in SSP phase is described by:

$$M(\theta)\ddot{\theta} + H(\theta, \dot{\theta}) + G(\theta) = DU \quad (3)$$

where $\ddot{\theta} = [\ddot{\theta}_1\ \ddot{\theta}_2\ \ddot{\theta}_3\ \ddot{\theta}_4\ \ddot{\theta}_5]^T$ is the vector of joint accelerations, $U = [U_1\ U_2\ U_3\ U_4\ U_5]^T$ is the vector of torque inputs, $M(\theta)$ is the symmetric positive definite inertia matrix, $H(\theta, \dot{\theta})$ is the vector of centripetal and Coriolis torques and $G(\theta)$ is the vector of gravitational torques given by

$$M(\theta) = [M_{ij}]_{\substack{1 \le i \le 5 \\ 1 \le j \le 5}}, \quad H(\theta, \dot{\theta}) = [h_{ij}]_{\substack{1 \le i \le 5 \\ 1 \le j \le 5}}, \quad G(\theta) = [G_i]_{1 \le i \le 5}$$

where:

$M_{11} = m_1 k_1^2 + I_1 + (m_2 + m_3 + m_4 + m_5)l_1^2$
$M_{12} = (m_2 l_1 k_2 + m_3 l_1 l_2 + m_4 l_1 l_2 + m_5 l_1 l_2)\cos(\theta_1 - \theta_2)$
$M_{13} = m_3 l_1 k_3 \cos(\theta_1 - \theta_3)$
$M_{14} = (m_4 l_1(l_4 - k_4)) + m_5 l_1 l_5 \cos(\theta_1 + \theta_5)$
$M_{15} = m_5 l_1 (l_5 - k_5) \cos(\theta_1 + \theta_5)$
$M_{21} = (m_2 l_1 k_2 + (m_3 + m_4 + m_5)l_1 l_2)\cos(\theta_1 - \theta_2)$
$M_{22} = m_2 k_2^2 + I_2 + (m_3 + m_4 + m_5)l_2^2$
$M_{23} = m_3 l_2 k_3 \cos(\theta_2 - \theta_3)$
$M_{24} = (m_4 l_2(l_4 - k_4) + m_5 l_2 l_4)\cos(\theta_2 + \theta_4)$
$M_{25} = m_5 l_2 (l_5 - k_5) \cos(\theta_2 + \theta_5)$

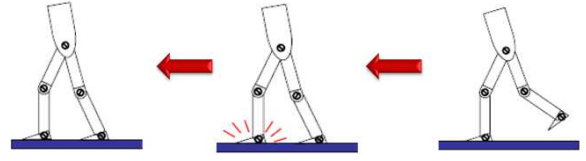

Figure 2. The three phases of a typical walking cycle

$M_{31} = m_3 l_1 k_3 \cos(\theta_1 - \theta_3)$
$M_{32} = m_3 l_2 k_3 \cos(\theta_2 - \theta_3)$
$M_{33} = m_3 k_3^2 + I_3$
$M_{34} = J_{35} = 0$
$M_{41} = (m_4 l_1(l_4 - k_4) + m_5 l_1 l_4)\cos(\theta_1 + \theta_4)$
$M_{42} = (m_4 l_2(l_4 - k_4) + m_5 l_2 l_4)\cos(\theta_2 + \theta_4)$
$M_{43} = 0$
$M_{44} = m_4(l_4 - k_4)^2 + m_5 l_4^2 + I_4$
$M_{45} = (m_5 l_4 (l_5 - k_5))\cos(\theta_4 - \theta_5)$
$M_{51} = m_5 l_1 (l_5 - k_5) \cos(\theta_1 + \theta_5)$
$M_{52} = m_5 l_2 (l_5 - k_5) \cos(\theta_2 + \theta_5)$
$M_{53} = 0$
$M_{54} = m_5 l_4 (l_5 - k_5) \cos(\theta_4 - \theta_5)$
$M_{55} = m_5 (l_5 - k_5)^2 + I_5$
$h_{11} = 0$
$h_{12} = (m_2 l_1 k_2 + m_3 l_1 l_2 + m_4 l_1 l_2 + m_5 l_1 l_2)\sin(\theta_1 - \theta_2)$
$h_{13} = (m_5 l_1 k_3)\sin(\theta_1 - \theta_3)$
$h_{14} = -(m_4 l_1(l_4 - k_4) + m_5 l_1 l_4)\sin(\theta_1 + \theta_4)$
$h_{15} = -(m_5 l_1 (l_5 - k_5))\sin(\theta_1 + \theta_5)$
$h_{21} = -(m_2 l_1 k_2 + (m_3 + m_4 + m_5)l_1 l_2)\sin(\theta_1 - \theta_2)$
$h_{22} = 0$
$h_{23} = m_3 l_2 k_3 \sin(\theta_2 - \theta_3)$
$h_{24} = -(m_4 l_2(l_4 - k_4) + m_5 l_2 l_4)\sin(\theta_2 + \theta_4)$
$h_{25} = -m_5 l_2 (l_5 - k_5)\sin(\theta_2 + \theta_5)$
$h_{31} = -m_3 l_1 k_3 \sin(\theta_1 - \theta_3)$
$h_{32} = -m_3 l_2 k_2 \sin(\theta_2 - \theta_3)$
$h_{33} = 0$
$h_{34} = h_{35} = 0$
$h_{41} = -(m_4 l_1(l_4 - k_4) + m_5 l_1 l_4)\sin(\theta_1 + \theta_4)$
$h_{42} = -(m_4 l_2(l_4 - k_4) + m_5 l_2 l_4)\sin(\theta_2 + \theta_4)$
$h_{43} = 0$
$h_{44} = 0$
$h_{45} = (m_5 l_4 (l_5 - k_5))\sin(\theta_4 - \theta_5)$
$h_{51} = -(m_5 l_1 (l_5 - k_5))\sin(\theta_1 + \theta_5)$
$h_{52} = -(m_5 l_2 (l_5 - k_5))\sin(\theta_2 + \theta_5)$
$h_{53} = h_{55} = 0$
$h_{54} = -(m_5 l_4 (l_5 - k_5))\sin(\theta_4 - \theta_5)$

$$G_1 = g(m_1 k_1 + (m_2 + m_3 + m_4 + m_5)l_1)\cos\theta_1$$
$$G_2 = g(m_2 k_2 + (m_3 + m_4 + m_5)l_2)\cos\theta_2$$
$$G_3 = m_3 g k_3 \cos\theta_3$$
$$G_4 = g(m_4(l_4 - k_4) + m_5 l_4)\cos\theta_4$$
$$G_5 = g(m_5(l_5 - k_5))\cos\theta_5$$
$$D = \begin{pmatrix} 1 & -1 & 0 & 0 & 0 \\ 0 & 1 & -1 & 0 & 0 \\ 0 & 0 & 1 & -1 & 0 \\ 0 & 0 & 0 & 1 & -1 \\ 0 & 0 & 0 & 0 & 1 \end{pmatrix}$$

## IV. MODEL REDUCTION AND GAIT TRAJECTORY GENERATION

The purpose of this section is to reduce the nonlinear dynamic model (3) into a solvable differential system in order to give algebraic solutions of gait trajectories. The reduced model will be obtained as follows: the robotic system (3) is first written around an equilibrium point $\theta_{eq}$ as [9]:

$$\dot{x} = A\,x + B\,v \quad (4)$$

$$A = \begin{bmatrix} 0_{5\times 5} & I_{5\times 5} \\ -J^{-1}\dfrac{\partial G}{\partial \theta}\Big|_{\theta_{eq}} & 0_{5\times 5} \end{bmatrix} \qquad B = \begin{bmatrix} 0_{5\times 5} \\ J^{-1}D \end{bmatrix}$$

$$v = \begin{bmatrix} 0_{1\times 5} & U - U_{eq} \end{bmatrix}^T$$

$$x = \begin{bmatrix} \theta - \theta_{eq} & \dot{\theta} - \dot{\theta}_{eq} \end{bmatrix}^T$$

To generate joint desired joint trajectory vector:

$$\theta_d(t) = \begin{bmatrix} \theta_{1,d}(t) & \theta_{2,d}(t) & \theta_{3,d}(t) & \theta_{4,d}(t) & \theta_{5,d}(t) \end{bmatrix}^T$$

we assume, for the bipedal robot, the following biologically inspired assumptions (see Fig.3):

*Assumption* 1: ZMP stability [10] is imposed. This is can be ensured by assuming that:
$$U_1 = 0 \quad (5)$$
*Assumption* 2: the supporting leg is kept straight as:
$$\theta_{1,d}(t) = \theta_{2,d}(t) \quad (6)$$
*Assumption* 3: The bipedal robot must have an upright posture that is to say, it keeps its back straight such that:
$$\theta_{3,d}(t) = \frac{\pi}{2} \quad (7)$$
*Assumption* 4: the relationship between the right ankle joint and the left hip joint is imposed such that:
$$\theta_{4,d}(t) = \theta_{1,d}(t) + \alpha \quad (8)$$
where $\alpha$ is a given constant angle to be chosen.
*Assumption* 5: To generate a rhythmic stable movement, the relationship between the right ankle joint and the left knee joint is given by [11]:
$$\theta_{5,d}(t) = \theta_{1,d}(t) - \sin^2(\frac{\pi}{T}t) \quad (9)$$

From the assumptions (6)-(9), we can then easily deduce the desired velocity vector and the desired acceleration vector defined, respectively, by:

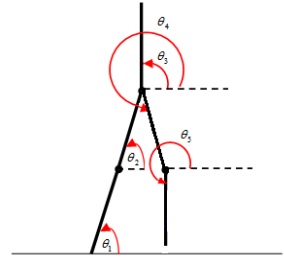

Figure 3. Biologically inspired assumptions

$$\dot{\theta}_d(t) = \begin{bmatrix} \dot{\theta}_{1,d}(t) & \dot{\theta}_{2,d}(t) & \dot{\theta}_{3,d}(t) & \dot{\theta}_{4,d}(t) & \dot{\theta}_{5,d}(t) \end{bmatrix}^T$$
$$\ddot{\theta}_d(t) = \begin{bmatrix} \ddot{\theta}_{1,d}(t) & \ddot{\theta}_{2,d}(t) & \ddot{\theta}_{3,d}(t) & \ddot{\theta}_{4,d}(t) & \ddot{\theta}_{5,d}(t) \end{bmatrix}^T$$

Particularly the following relations are deduced:
$$\ddot{\theta}_{1,d}(t) = \ddot{\theta}_{2,d}(t) \quad (10)$$
$$\ddot{\theta}_{3,d}(t) = 0 \quad (11)$$
$$\ddot{\theta}_{4,d}(t) = \ddot{\theta}_{1,d}(t) \quad (12)$$
$$\ddot{\theta}_{5,d} = \ddot{\theta}_{1,d} + 2\left(\frac{\pi}{T}\right)^2 \cos\left(2\frac{\pi}{T}t\right) \quad (13)$$

By summing all the lines of the system (4) and using the relation (5), (10)-(13), the following reduced homogeneous 2nd order differential equations with constant coefficients is obtained as:

$$M_1 \ddot{\theta}_{1,d}(t) + H_1 \theta_{1,d}(t) = -K - h(t) - s(t) \quad (14)$$

where:
$$M_1 = (M_{11} + M_{22} + M_{44} + M_{55} + M_{23})\big|_{\theta_{eq}}$$
$$\qquad + 2(M_{12} + M_{14} + M_{15} + M_{25} + M_{45})\big|_{\theta_{eq}}$$

$$H_1 = \left(\frac{\partial G_1}{\partial \theta_1} + \frac{\partial G_2}{\partial \theta_2} + \frac{\partial G_4}{\partial \theta_4} + \frac{\partial G_5}{\partial \theta_5}\right)\Bigg|_{\theta_{eq}}$$

$$K = -\left(\frac{\partial G_1}{\partial \theta_1}\theta_1 + \frac{\partial G_2}{\partial \theta_2}\theta_2 + \frac{\partial G_3}{\partial \theta_3}\theta_3 + \frac{\partial G_4}{\partial \theta_4}\theta_4 + \frac{\partial G_5}{\partial \theta_5}\theta_5\right)\Bigg|_{\theta_{eq}}$$

$$\qquad + \left(\alpha \frac{\partial G_4}{\partial \theta_4} + \frac{\pi}{2}\frac{\partial G_3}{\partial \theta_3} + \frac{1}{2}\frac{\partial G_5}{\partial \theta_5}\right)\Bigg|_{\theta_{eq}} + U_{1eq}$$

$$h(t) = -\frac{\partial G_5}{2\partial \theta_5}\left(\cos 2\frac{\pi}{T}t\right)$$

$$s(t) = 2\left(\frac{\pi}{T}\right)^2 M_2 \cos\left(2\frac{\pi}{T}t\right)$$

$$M_2 = M_{15} + M_{25} + M_{45} + M_{55}$$

Solving (14) for the boundary conditions:
$$\theta_{1,d}(t_0) = \theta_{10}$$
$$\theta_{1,d}(t_f) = \theta_{1f}$$

and for the equilibrium point :

$$\theta_{eq} = \begin{bmatrix} \frac{\pi}{2} & \frac{\pi}{2} & \frac{\pi}{2} & \frac{\pi}{2} & \frac{\pi}{2} \end{bmatrix}^T$$

the analytical solution of the differential equation (14) is given by [12]:

$$\theta_{1,d}(t) = C_1 e^{r_1 t} + C_2 e^{r_2 t} + C_3 \cos\left(2\frac{\pi}{T}t\right) - \frac{K}{H_1} \quad (15)$$

where :

$$C_1 = C_2 - \frac{K}{H_1} + C_3 - \theta_{10}$$

$$C_2 = \frac{\left(C_3 + \theta_{1f} - \theta_{10} - \frac{K}{H_1}\right)e^{r_1 t_f} - C_3 \cos\left(2\frac{\pi}{T}t_f\right) + K/H_1}{e^{r_1 t_f} + e^{r_2 t_f}}$$

$$C_3 = \frac{\frac{\partial G_5}{2\partial \theta_5} - M_2\left(\frac{\pi}{T}\right)^2}{\left(4\left(\frac{\pi}{T}\right)^2 M_1 - H_1\right)}$$

$$r_{1,2} = \pm \frac{\sqrt{-M_1 \cdot H_1}}{M_1}, \quad M_1 H_1 \langle 0$$

To verify the algebraic solution (15) that asked a rather tedious calculation two approaches are used: a symbolic computation of the solution (15) using the symbolic toolbox of Matlab software and a numerical integration of the differential equation (14) using the ode45 function of the same software. The three solutions were found superimposed for the physical parameters given by Table 1.

Using the relations (15) and (6)-(9), the analytical expressions of the joint trajectories are then deduced. We impose then to the robotic model (3) the following second order linear input-output behavior [8]:

$$\left(\ddot{\theta}(t) - \ddot{\theta}_d(t)\right) + K_v\left(\dot{\theta}(t) - \dot{\theta}_d(t)\right) + K_p\left(\theta(t) - \theta_d(t)\right) = 0 \quad (16)$$

where $K_v \in \Re^{5 \times 5}$ and $K_p \in \Re^{5 \times 5}$ are two positive definite diagonal matrices chosen to guarantee global stability, desired performances and decoupling proprieties for the controlled system and such that the desired trajectories $\theta_d$, $\dot{\theta}_d$ and are derived from the solution (15) and the relations (6)-(9). The control law deduced from (3) and (16) is then given by:

$$U = D^{-1}\bigl[M(\theta)\bigl(\ddot{\theta}_d(t) - K_v\bigl(\dot{\theta}(t) - \dot{\theta}_d(t)\bigr) - K_p\bigl(\theta(t) - \theta_d(t)\bigr)\bigr) \\ + H(\theta,\dot{\theta}) + G(\theta)\bigr] \quad (17)$$

Using the physical parameters given in table1, the joint trajectories of the controlled biped robot are generated as see by Fig. 4 where the desired joint position is designed by $\theta_d(t) = [teta_1(t) \ teta_2(t) \ tata_3(t) \ teta_4(t) \ teta_5(t)]^T$. The walking cycle is shown by Fig.5.

Compared to previous works [8], this paper gives an algebraic solution for desired joint trajectories that must be followed by the robotic system. The result is a stable rhythmic movement for the planar bipedal system.

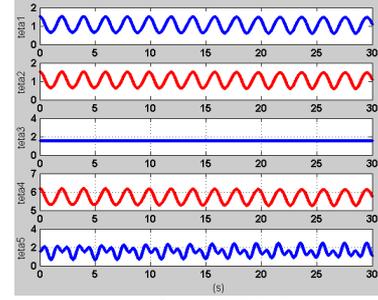

Figure 4. Joint trajectories of the bipedal robot in the swing phase

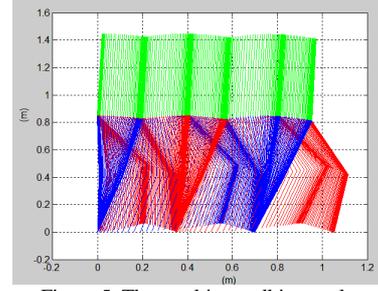

Figure 5. The resulting walking cycle

## V. CONCLUSION

A simple trajectory generation method is proposed for bipedal gait by imposing a sinusoidal curve to the ankle of the swing leg's trajectory.